\documentclass[pdflatex,sn-mathphys-num]{sn-jnl}% Math and Physical Sciences Numbered Reference Style 
\usepackage{graphicx}%
\usepackage{multirow}%
\usepackage{amsmath,amssymb,amsfonts}%
\usepackage{amsthm}%
\usepackage{mathrsfs}%
\usepackage[title]{appendix}%
\usepackage{xcolor}%
\usepackage{textcomp}%
\usepackage{manyfoot}%
\usepackage{booktabs}%
\usepackage{listings}%

\usepackage{algorithm}%
\usepackage{algorithmicx}%
\usepackage{algpseudocode}%

\usepackage{subcaption}% For subcaption in figures.

\usepackage{tikz}
\usepackage{amsmath} 
\usetikzlibrary{calc, positioning, shapes, fit}
\usetikzlibrary{arrows.meta, calc, decorations.pathreplacing,
                ext.paths.ortho, positioning, quotes}
\usepackage{tikzpeople}
\makeatletter
\tikzset{phantom node/.code=\tikz@addoption{\expandafter\let\csname pgf@sh@boxes@\tikz@shape\endcsname\pgfutil@empty}}
\makeatother
\tikzset{
  shadowed node xshift/.initial=1.5ex, shadowed node yshift/.initial=1ex, shadowed node list/.initial={2, 1},
  pics/shadowed node/.default=\pgfkeysvalueof{/tikz/shadowed node list},
  shadowed node/.pic={
    \foreach[expand list] \elem in {#1}
      \scoped[transparency group, shadowed node calculation={\elem}]
        \node[style/.expand once=\tikzpictextoptions, phantom node,
              xshift={-\elem*\pgfkeysvalueof{/tikz/shadowed node xshift}}, % changed xshift to negative to reverse direction
              yshift={\elem*\pgfkeysvalueof{/tikz/shadowed node yshift}}] (-\elem) {\tikzpictext};
    \node[alias=-0, style/.expand once=\tikzpictextoptions] () {\tikzpictext};},
  set shadowed node calculation parameter/.style={shadowed node calculation/.style={opacity={(#1-##1+1)/(#1+1)}}},
  set shadowed node calculation parameter=2,
  overshoot line to/.style={to path={($(\tikztostart)!-(#1)!(\tikztotarget)$)--($(\tikztotarget)!-(#1)!(\tikztostart)$)\tikztonodes}},
  edges have transparency group/.style={execute at begin to={\scope[transparency group,#1]}, execute at end to=\endscope}}

\theoremstyle{thmstyleone}%
\theoremstyle{thmstyletwo}%

\theoremstyle{thmstylethree}%

\begin{document}

\title[Article Title]{DUAL: Diversity and Uncertainty Active Learning for Text Summarization}


\author*[1]{\fnm{Petros Stylianos} \sur{Giouroukis}}\email{pgiouroukis@gmail.com}

\author[1]{\fnm{Alexios} \sur{Gidiotis}}\email{alexgidiotis@gmail.com}

\author[1]{\fnm{Grigorios} \sur{Tsoumakas}}\email{greg@csd.auth.gr}

\affil[1]{\orgdiv{School of Informatics}, \orgname{Aristotle University of Thessaloniki}, \orgaddress{\street{Aristotle University Campus}, \city{Thessaloniki}, \postcode{54124}, \country{Greece}}}


\abstract{With the rise of large language models, neural text summarization has advanced significantly in recent years. However, even state-of-the-art models continue to rely heavily on high-quality human-annotated data for training and evaluation. Active learning is frequently used as an effective way to collect such datasets, especially when annotation resources are scarce. Active learning methods typically prioritize either uncertainty or diversity but have shown limited effectiveness in summarization, often being outperformed by random sampling. We present Diversity and Uncertainty Active Learning (DUAL), a novel algorithm that combines uncertainty and diversity to iteratively select and annotate samples that are both representative of the data distribution and challenging for the current model. DUAL addresses the selection of noisy samples in uncertainty-based methods and the limited exploration scope of diversity-based methods. Through extensive experiments with different summarization models and benchmark datasets, we demonstrate that DUAL consistently matches or outperforms the best performing strategies. %, maintaining the reliability of random sampling as a fallback. 
Using visualizations and quantitative metrics, we provide valuable insights into the effectiveness and robustness of different active learning strategies, in an attempt to understand why these strategies haven't performed consistently in text summarization. Finally, we show that DUAL strikes a good balance between diversity and robustness.}

\keywords{Active learning, Abstractive text summarization, Natural language processing, Deep learning, Large language models}



\maketitle

\section{Introduction}
\label{sec:introduction}

Neural text summarization methods have achieved impressive performance in recent years \cite{Lewis2019BART:Comprehension, Zhang2020PEGASUS:Summarization, Guo2022LongT5:Sequences, Xiong2023AdaptingSequences}, driven by the great advances in Natural Language Processing (NLP). Large Language Models (LLMs), based on the Transformer architecture \cite{Vaswani2017AttentionNeed}, have become the predominant approach for solving any Natural Language Generation (NLG) problem \cite{Lewis2019BART:Comprehension, liu2019roberta, Raffel2020ExploringTransformer}, including summarization. These models rely on extensive pre-training on massive amounts of text data, in order to generalize to a variety of downstream tasks \cite{Radford2018ImprovingPre-training, Lewis2019BART:Comprehension, Radford2019LanguageLearners, Raffel2020ExploringTransformer}. Nonetheless, in most cases, they still require some task specific fine-tuning. Contrary to previous generation non pre-trained neural models, state-of-the-art summarization models no longer require thousands of labeled training data points, and can achieve great performance with much fewer samples. A consequence of this shift to smaller datasets is that the quality of each training sample is now even more important. The impact of even a few noisy or non-representative training samples has a much greater effect on the final model \cite{Tsvigun2022ActiveSummarization, Gidiotis2023BayesianSummarization}. This amplifies the need for high quality annotated training data.

Active Learning (AL), has for many years been an obvious choice for collecting a small but high quality training set, especially for shallow learning models. Nevertheless, the application of AL methods to generative tasks such as summarization has been non-trivial \cite{Gidiotis2023BayesianSummarization}. The primary reason for this is that selecting highly informative and diverse samples for annotation is a very hard problem to solve. There's no obvious answer to what constitutes a good sample \cite{Perlitz2023ActiveGeneration}, and various classic AL strategies have failed to produce consistent results.

This work takes a step forward in addressing the challenge of selecting highly informative and diverse samples for summarization training. We explore the advantages and shortcomings of traditional uncertainty and diversity based strategies, and aim to understand why both of those approaches haven't been very effective in text summarization. Finally, we present Diversity and Uncertainty Active Learning (DUAL), a novel AL algorithm that combines both strategies to leverage their complementary strengths. Through extensive experiments on several well known models and benchmark datasets, we show that indeed this combination of methods produces significant improvements over existing strategies. Furthermore, the performance of DUAL is consistent across all models and datasets in our experiments.

The remainder of this article is organized as follows. Section \ref{sec:related} presents an overview of AL methods for NLP and summarization, emphasizing two complementary approaches closely related to our proposed method. Section \ref{sec:methods} details the methodology behind our approach. Section \ref{sec:experiments} describes our experimental setup. Section \ref{sec:results} discusses the experimental results and key findings. Finally, Section \ref{sec:conclusion} summarizes our contributions and findings, offering insights and potential directions for future research. The source code used in our experiments is publicly available online\footnote{\href{https://github.com/pgiouroukis/dual}{https://github.com/pgiouroukis/dual}}.

\section{Related Work}\label{sec:related}

%\subsection{Active Learning in NLP}
Active learning has been applied to a wide variety of NLP tasks. Firstly, in text classification, \cite{Ein-Dor2020ActiveStudy,Schroder2022RevisitingTransformers,Margatina2021ActiveExamples,Wu2022Context-awareLabeling,Yu2022ACTUNE:Models} have shown interesting results by applying different AL methods. Secondly, in Neural Machine Translation (NMT), AL has been particularly useful in low-resource language pairs \cite{Haffari2009ActiveTranslation,Zheng2019SentenceSummarization,Zhao2020ActiveTranslation, Liu2023Uncertainty-awareTranslation}. Finally, in named entity recognition, the application of AL methods has also resulted in performance improvements \cite{Shen2018DeepRecognition, Siddhant2020DeepStudy,
Radmard2021SubsequenceRecognition, Liu2020LTP:Recognition}.

In contrast, for most generative tasks other than NMT, AL methods have struggled to produce consistent results, often being outperformed by simple baselines such as random sampling \cite{Tsvigun2022ActiveSummarization,Gidiotis2023BayesianSummarization}. \cite{Perlitz2023ActiveGeneration} evaluates existing AL strategies in NLG tasks such as paraphrase generation, summarization and question generation. Interestingly, the authors suggest that compared to classification, NLG tasks lack a clearly defined ground-truth, which contributes to the poor performance of AL methods.

Most AL methods in NLG, and summarization in particular, focus on one of two strategies: uncertainty sampling and diversity sampling. In the following two sections, we delve deeper into the AL literature for text summarization, examining two complementary methods, one based on uncertainty and the other on diversity. These methods form the foundation of our work, so we will present each one in detail.

% \cite{Karaoguz
% (2018)} applies AL to paraphrase generation with LSTMs,
% Finally, in summarization, AL has seen limited adoption. \cite{Zhang2009ActiveSummarization} propose an active learning method that selects unlabeled training documents based on their similarity with PowerPoint slides from corresponding presentations.

\subsection{Bayesian Active Summarization}
Bayesian Active Summarization (BAS) \cite{Gidiotis2023BayesianSummarization} applies a Bayesian deep learning framework to quantify summarization uncertainty, and uses it as query strategy to select data instances. In particular, BAS uses Monte Carlo (MC) dropout \cite{Gal2016DropoutLearning} in order to efficiently approximate sampling from the predictive distribution of a neural summarization model. In practice, this involves generating $N$ summaries for the same input with dropout enabled. The differences between these summaries $y_i$ and $y_j$, are measured using the BLEU metric \cite{Papineni2002BLEU:Translation} $\text{BLEU}(y_i,y_j)$. Summarization uncertainty is then estimated by calculating the BLEU Variance (BLEUVar) \cite{Gidiotis2022ShouldRescue} across all $N$ summaries, as described in Equation \ref{eq:bleuvar}.

\begin{equation}
\label{eq:bleuvar}
\text{BLEUVar} = \frac{1}{N(N-1)}\sum_{i=1}^{N} \sum_{j \neq i}^{N} (1 - \text{BLEU}(y_i,y_j))^2
\end{equation}

At each iteration of BAS a subset of the unlabeled data instances is scored using a summarization model with MC dropout, and BLEUVar is computed. Then the samples of this subset with the highest BLEUVar are selected for annotation, added to the labeled set and the summarization model is trained on the extended labeled set. This method has shown promising results with very few training instances on several benchmark datasets using the BART \cite{Lewis2019BART:Comprehension} and PEGASUS \cite{Zhang2020PEGASUS:Summarization} summarization models.

\subsection{In-Domain Diversity Sampling}
% [: if we have a section where we discuss uncertainty vs diversity, then this paragraph should move there]
% In contrast to uncertainty strategies, representativeness-based strategies aim to select a diverse and representative set of examples for labeling. Given a training set of labeled samples $L$ and a pool of unlabeled samples $U$, different strategies are used to select samples that will result in a training set representative of the real distribution.

In-Domain Diversity Sampling (IDDS) \cite{Tsvigun2022ActiveSummarization} ranks unlabeled samples based on their similarity to the unlabeled set and their dissimilarity to the labeled set. This encourages the selection of samples that are representative of the unlabeled data while being dissimilar from already labeled instances. Formally, for each sample in the pool of unlabeled documents, IDDS computes a score as follows:

\begin{equation}
\label{eq:idds}
\text{IDDS}(x) = \lambda \frac{ \sum_{j=1}^{|U|} s(\phi(x), \phi(x_j))}{|U|}
- (1 - \lambda) \frac{\sum_{i=1}^{|L|} s(\phi(x), \phi(x_i))}{|L|}
\end{equation}
where $\phi(x)$ denotes the embedding of document $x$, $s(i, j)$ is the similarity between $i$ and $j$, and $U$ and $L$ are the unlabeled and labeled data pools respectively. In the equation above, $\lambda$ is a balancing parameter with a value between 0 and 1. 

In order to compute the similarity scores, IDDS requires the calculation of vector representations for all labeled and unlabeled samples. At each step, the IDDS scores are determined for all documents in $U$, and a sample $x^*$ that maximizes the IDDS score is selected and moved from $U$ to $L$. This process repeats, recalculating the scores for the remaining documents in $U$, until the annotation budget $B$ is reached. Importantly, the model is not involved in the selection process, allowing $B$ samples to be selected without intermediate model training steps.

\section{Methodology}\label{sec:methods}

% While BAS, IDDS and other AL strategies demonstrate strong results in specific datasets and with specific models, there is no strategy that consistently outperforms random sampling. IDDS may fail to identify challenging or highly informative samples, as it focuses primarily on maintaining diversity. On the other hand, BAS may select noisy samples due to high uncertainty, which can negatively impact model performance. Furthermore, as shown in, random sampling has been found to be unexpectedly effective and is often challenging to outperform in AL settings. 
% We argue that IDDS can become confined within a limited neighborhood of the embedding space, which may not provide a sufficiently representative view of the overall dataset. This limitation is compounded by the inherent challenge of generating high-quality embeddings that fully capture the nuances of the data. These factors emphasize the necessity of incorporating random samples into our selection strategy to ensure a more comprehensive and balanced approach to active learning.

AL methods have mainly focused either on uncertainty-based or diversity-based sample selection strategies. While both types of strategies have a solid theoretical background, they also have different shortcomings. Pure uncertainty-based strategies can be problematic because noisy samples tend to have high uncertainty, increasing their likelihood of being selected, and can be harmful when a model is trained on them \cite{Tsvigun2022ActiveSummarization, Gidiotis2023BayesianSummarization}. On the other hand, as we show in this work, diversity-based strategies can become confined within a limited region of the embedding space, which in turn might lead to a sample that's not representative of the entire dataset. This limitation is compounded by the inherent challenge of generating high-quality embeddings that fully capture all the nuances of the data.

To overcome these limitations, our approach combines both uncertainty- and diversity-based queries, leveraging their strengths while mitigating their shortcomings. The outcome is Diversity and Uncertainty Active Learning (DUAL), a method that iteratively selects and annotates the most diverse and informative data samples, in order to arrive at a better trained model. 

We start with a pre-trained model $M$, a pool of unlabeled samples $U$ and an initially empty labeled pool $L$.~DUAL iteratively selects $s$ samples using a strategy that integrates both diversity-based and uncertainty-based selection, until the annotation budget $B$ is exhausted. At each iteration, we first compute the IDDS score for all document vectors $\phi(x) \in U$, using Equation \ref{eq:idds}. We then select the top-$k$ samples with the highest scores, creating a subset $K \subset U$. The selected samples form a relatively tight neighborhood in the embedding space, as a natural consequence of the IDDS scoring mechanism. This diversity-based selection step aims to identify regions of the embedding space were samples are highly representative of the data distribution yet underepresented in the labeled set.

After selecting the diverse set $K$, we estimate the uncertainty of all $k$ samples using the BAS method. For each sample $x \in K$, we perform $n$ stochastic forward passes with dropout enabled in $M$. Using Equation \ref{eq:bleuvar}, we compute the BLEUVar score for each $x$ as an uncertainty metric. Following \cite{Gidiotis2023BayesianSummarization}, to avoid selecting noisy samples that could harm performance, we exclude samples with BLEUVar scores exceeding a threshold $\tau$. From the remaining candidates, we select the sample with the highest BLEUVar score moving it from $U$ to $L$, while removing the rest of the $K$ set from the pool. This selection process is repeated, starting again from the IDDS scoring step, until the required number of samples is selected.

% Random sampling
As shown in \cite{Gidiotis2023BayesianSummarization}, including a percentage of randomly selected samples benefits performance. Our method introduces a parameter $p$ between 0 and 1, that controls the amount of random samples we select at each AL iteration. In order to reach the required $s$ samples, we first select $s_1 = p \cdot s$ samples using the DUAL strategy, and then we select the remaining $s_2 = (1-p) \cdot s$ samples by randomly sampling from the unlabeled pool $U$. This random sampling step is crucial for several reasons. First, it mitigates the tendency of IDDS to focus on limited regions, thereby enhancing diversity. Second, it encourages exploration in the AL process.
% Finally, we found that with the appropriate value of $p$ our method performs at least as well as the random strategy, even if the targeted approach underperforms in the same setting.

\begin{figure}[h]
\label{fig:proposed_method}
  \centering
  \resizebox{\textwidth}{!}{	\begin{tikzpicture}[
            thick,
            n/.style={rounded corners, draw, fill={#1!20}},
            > = {Stealth[round, sep]}]

                \node[n = green, align=center] (unlabeled-data) at (0,0) {Unlabeled \\ Data};

                \node[n = yellow, align=center, fill=orange!30!white, top color=orange!0, bottom color=orange!70,] (compute-idds) at ($(unlabeled-data) + (3cm,2.85cm)$) {Compute \\ IDDS Scores};

                % \node[n = red, align=center] (labeled-data) [below=2cm of compute-idds] {Labeled \\ Data};

                \node[cylinder,
                draw,
                align = center,
        	cylinder uses custom fill, 
        	cylinder body fill = green!10, 
        	cylinder end fill = green!20,
        	aspect = 0.2, 
                minimum width = 1.4cm,
	        minimum height = 1.2cm,
        	shape border rotate = 90] (labeled-data) [below=1.75cm of compute-idds]         {Labeled \\ Data};      
            
                \pic["Document" n=blue] (documents) [right=1.9cm of compute-idds] {shadowed node};
                \draw[
                arrows={[arc=135]}, arrows={Hooks[left]-Hooks[right]},
                s/.style={shift={(.75mm,0.4mm)}}]
                ([s]documents-2.north) to[overshoot line to=1mm] coordinate(@) ([s]documents.north);
                \path[every pin edge/.style={black}] node also[pin=above:$k$](@);

                 \node[n = red, fill=red!30!white, top color=red!10, bottom color=red!50, inner sep=10pt] (model) at ($(documents) + (1.5cm,-2.95cm)$) {Model};

		      % \pic["Summary" n=blue] (summary-mid) [right=2cm of documents] {shadowed node};
                \node(summary-mid) [right=3.5cm of documents] {\vdots}; 

		      \pic["Summary" n=blue] (summary-top) [above=0cm of summary-mid] {shadowed node};
                \draw[
                arrows={[arc=135]}, arrows={Hooks[left]-Hooks[right]},
                s/.style={shift={(.75mm,0.4mm)}}]
                ([s]summary-top-2.north) to[overshoot line to=1mm] coordinate(@) ([s]summary-top.north);
                \path[every pin edge/.style={black}] node also[pin=above:$n$](@);  
                
                \pic["Summary" n=blue] (summary-bottom) [below=0.3cm of summary-mid] {shadowed node};

                \draw[decorate,decoration={brace,amplitude=15pt,mirror}] 
  ([xshift=-0.2cm, yshift=0.3cm]summary-top.north west) -- ([xshift=-0.2cm]summary-bottom.south west) 
  node[midway,xshift=-0.8cm] {$k$};

		      \pic["BLEUVar" n=blue] (bleuvar) [right=1.5cm of summary-mid] {shadowed node};
                \draw[
                arrows={[arc=135]}, arrows={Hooks[left]-Hooks[right]},
                s/.style={shift={(.75mm,0.4mm)}}]
                ([s]bleuvar-2.north) to[overshoot line to=1mm] coordinate(@) ([s]bleuvar.north);
                \path[every pin edge/.style={black}] node also[pin=above:$k$](@); 

                \node[n = orange, align=center, fill=orange!30!white, top color=orange!0, bottom color=orange!70,] (filter-argmax) at ($(bleuvar) + (2.75cm,0.75cm)$) {Filter-Argmax};

                \node[n = blue, align=center] (document) at ($(bleuvar) + (2.75cm,-0.75cm)$) {Document};

                \node[bob,mirrored,minimum size=1.2cm] (person) [below=2cm of document] {};

                \node[n = yellow, align=center, fill=orange!30!white, top color=orange!0, bottom color=orange!70] (random-sample) [below=2cm of labeled-data] {Random \\ Sample};

                \node[n = blue, align=center] (document2) [right=3cm of random-sample] {Document};

                \draw[] (model.east) -- ([xshift=-0.73cm, yshift=0.15cm]$(summary-top.north west)!0.5!(summary-bottom.south west)$);

                \draw[-stealth] (unlabeled-data.east) -- (compute-idds.west) node[midway, left] {$s_1$ samples};
                \draw[-stealth] (labeled-data.north) -- (compute-idds.south);
                \draw[-stealth] (labeled-data.north) -- (compute-idds.south);
                \draw[-stealth] (compute-idds.east) -- ($(compute-idds.east)!0.85!(documents.west)$) node[midway, below, align=center] {select \\ $top_k$};
                \draw[-stealth] (documents.south) -- (model.west);
                % \draw[-stealth] (model.north) -- (model.west);
                % \draw[-stealth] (model.east) -- ++([rotate=7, scale=1.05]$(summary-top.west) - (model.east)$);
                % \draw[-stealth] (model.east) -- ++([rotate=4, scale=1]$(summary-mid.west) - (model.east)$);
                % \draw[-stealth] (model.east) -- ++([rotate=-3, scale=0.93]$(summary-bottom.west) - (model.east)$);                
                \draw[-stealth] (summary-top.east) -- (bleuvar.west);
                % \draw[-stealth] (summary-mid.east) -- (bleuvar.west);
                \draw[-stealth] (summary-bottom.east) -- (bleuvar.west);
                \draw[-stealth] (bleuvar.east) -- (filter-argmax.west);
                \draw[-stealth] (filter-argmax.south) -- (document.north);
                \draw[-stealth] (document.south) -- (person.north) node[midway, left] {annotate};
                \draw[->] (person.west) to[out=170, in=350] (labeled-data.south);
                \draw[-stealth] (labeled-data.east) -- (model.west) node[midway, above, align=center] {finetune after $s$ \\ collected samples};
                \draw[-stealth] (unlabeled-data.east) -- (random-sample.west) node[midway, left] {$s_2$ samples};
                \draw[-stealth] (random-sample.east) -- (document2.west);
                \draw[-stealth] (document2.east) -- (person.south) node[midway, right] {annotate};

                \node[draw,
                rounded corners=0.3cm,
                fill=yellow!10,
                fill opacity=0.3,
                inner sep=0.5cm,    % Padding for the sides
                inner ysep=0.75cm,   % Padding for bottom
                yshift=0.5cm,      % Extend the top by shifting the fit box down
                fit={(compute-idds) (documents) (summary-mid) (summary-top) (summary-bottom) (bleuvar) (filter-argmax) (document)}, 
                name=fitbox] {};
                \node[anchor=north west, xshift=0.4cm, yshift=-0.4cm, align=left] at (fitbox.north west) {\textbf{Targeted Sampling Step} \\ repeats $s_1$ times};

                \node[draw,
                rounded corners=0.3cm,
                fill=yellow!10,
                fill opacity=0.3,
                inner sep=0.5cm,    % Padding for the sides
                inner ysep=0.75cm,   % Padding for bottom
                yshift=0.5cm,      % Extend the top by shifting the fit box down
                fit={(random-sample) (document2)}, 
                name=fitbox2] {};
                \node[anchor=north, yshift=-0.3cm, align=center] at (fitbox2.north) {\textbf{Random Sampling Step } \\ repeats $s_2$ times};

                \node[cylinder,
                draw,
                align = center,
        	cylinder uses custom fill, 
        	cylinder body fill = gray!10, 
        	cylinder end fill = gray!25,
        	aspect = 0.2, 
                minimum width = 1.8cm,
	        minimum height = 1.7cm,
        	shape border rotate = 90] (c) at (0,0) {Unlabeled \\ Data};

	\end{tikzpicture}}
  \caption{An iteration cycle of DUAL, comprising the selection of samples, their labeling and model retraining}
  \label{fig:dual}
\end{figure}
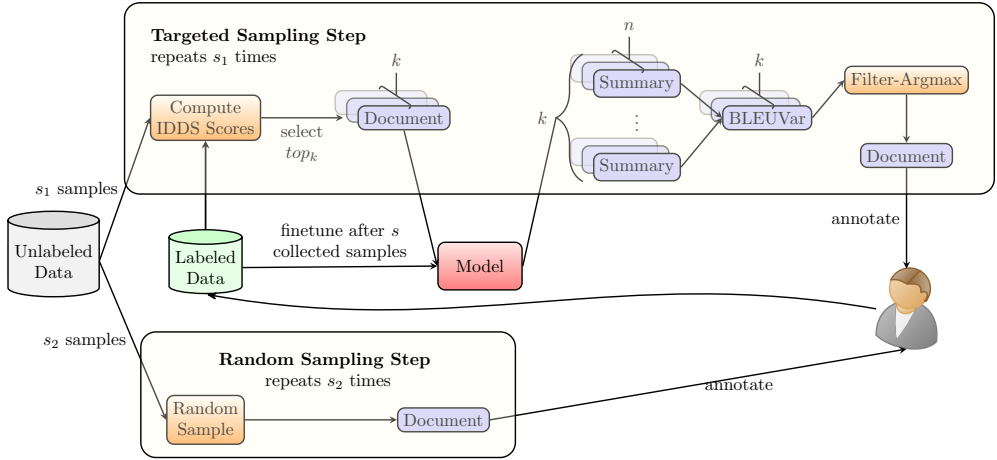

% End of iteration and retraining
After selecting all $s$ samples, we retrain model $M$ to incorporate the new labeled data, ensuring future uncertainty calculations reflect the updated model state. This cycle of selection, labeling, and retraining constitutes one iteration of our active learning approach, and is repeated until the annotation budget $B$ is exhausted. A full iteration of DUAL is shown in Figure \ref{fig:dual}.

% Details after the main method
Before the first active learning iteration we warm-start the model $M$ by training on a small set of $s_0$ labeled documents. This step is not necessary for modern LLMs, as they're already able to generalize to the summarization problem \cite{Radford2019LanguageLearners,Perlitz2023ActiveGeneration}. Nevertheless, we found that it helps make the first iteration more effective, as it adapts the model to the actual data distribution. The warm-up samples can be selected via random sampling or IDDS, as these methods do not depend on the model's uncertainty assessments.

% At each iteration of DUAL we end up labeling only one of the $K$ samples, but we remove all $k$ samples from the unlabeled pool, so they can't be selected in subsequent iterations. This addresses a specific challenge we encountered during our experiments. The challenge is that for certain dense regions of the embedding space, samples that lie very close are likely to be selected in different AL iterations. As a result, we end up oversampling a certain region of the space, which can lead to creating an imbalanced training set. By removing all $k$ samples from the pool, we ensure the algorithm consistently explores new regions of the embedding space, instead of focusing on a single dense region.

% Attempted a different version of the previous paragraph, in order to add the notation $E$ of the excluded set, which will help with the clarity of the Algorithm and Tikz figure.
% At each iteration of DUAL, we label only one of the samples in $K$, but the remaining $k-1$ samples are also removed from $U$ and added to an excluded set $E$, ensuring that these samples will not be selected in subsequent iterations. 
At each iteration of DUAL, we label one of the samples in $K$, but we move the remaining $k-1$ samples from $U$ to an excluded set $E$, ensuring that we cannot select these samples in subsequent iterations. This addresses a specific challenge we encountered during our experiments. More specifically, in certain dense regions of the embedding space, samples that lie very close to each other may be selected in different AL iterations. As a result, we end up oversampling a certain region of the space, which can lead to creating an imbalanced training set. When computing the IDDS scores, we consider $L \cup E$ alongside $L$. By removing all $k$ samples from the pool, we ensure the algorithm consistently explores new regions of the embedding space, instead of focusing on a single dense region. The pseudocode of the proposed method is provided in Algorithm \ref{alg:dual_algorithm}.

\begin{algorithm}[h]
\caption{DUAL Strategy}\label{alg:dual_algorithm}
\begin{algorithmic}[1]
\Require Pre-trained model \( M \), unlabeled pool \( U \), annotation budget \( B \), samples per iteration \( s \), warm-up size \( s_0 \), random ratio \( p \), uncertainty threshold \( \tau \), MC dropout passes \( n \)
\State \( L \leftarrow \emptyset \)
\State \( E \leftarrow \emptyset \)
\State \( s_1 \leftarrow \lceil p \cdot s \rceil \)
\State \( s_2 \leftarrow s - s_1 \)
\State \( W \leftarrow \text{select\_random}(U, s_0) \) \Comment{Optional warm-start}
\State \( L \leftarrow L \cup W \)
\State \( U \leftarrow U \setminus W \)
\State \( M_\theta \leftarrow \text{train}(M_\theta, L) \)
\While{\( |L| < B \)}
    \For{\( i \leftarrow 1 \) to \( s_1 \)} \Comment{Targeted Sampling Step} 
        \State \( \text{idds\_scores} \leftarrow \text{compute\_idds}(U, L \cup E) \)
        \State \( K \leftarrow \text{select\_top\_k}(k, U, \text{idds\_scores}) \)
        \ForAll{\( x \in K \)}
            \State \( \text{outputs} \leftarrow \text{generate\_MC\_dropout\_summaries}(x, n, M_\theta) \)
            \State \( \text{bleuvars}[x] \leftarrow \text{compute\_bleuvar}(\text{outputs}) \)
        \EndFor
        \State \( K \leftarrow \{ x \in K \mid \text{bleuvars}[x] \leq \tau \} \)
        \If{\( K = \emptyset \)}
            \State \textbf{break}
        \EndIf
        \State \( x^* \leftarrow \arg\max_{x \in K} \text{bleuvars}[x] \)
        \State \( L \leftarrow L \cup \{ x^* \} \)
        \State \( U \leftarrow U \setminus K \)
        \State \( E \leftarrow E \cup (K \setminus \{ x^* \}) \)
    \EndFor
    \State \( R \leftarrow \text{select\_random}(U, s_2) \) \Comment{Random Sampling Step}
    \State \( L \leftarrow L \cup R \)
    \State \( U \leftarrow U \setminus R \)
    \State \( M_\theta \leftarrow \text{train}(M_\theta, L) \)
\EndWhile
\end{algorithmic}
\end{algorithm}

It is important to note that while the IDDS process selects samples from a tight region in the embedding space, the uncertainty of these samples can vary significantly. This happens because uncertainty in summarization is not solely a function of a document's content or position in the embedding space, but also depends on the model's current state. Two documents that are very similar in content and style can pose different challenges to the summarization model. This variation in uncertainty within a tight neighborhood is a key aspect that our method exploits. In short, by selecting the sample with the highest BLEUVar score from the retrieved documents we achieve the following. First, we ensure that it is representative, as it was selected by IDDS. Second, because the model has high uncertainty for this sample, it is more likely to benefit performance if the model is trained on it.

\section{Experimental Setup}\label{sec:experiments}
This section presents the details of our experimental study. We start off by presenting the key hyper-parameters of our AL experimental framework, and then we introduce the models and datasets that were used in our experiments. Finally, we discuss the embeddings, as they are a core part of our method.

\subsection{Active Learning Framework}

In accordance with prior studies \cite{Gidiotis2023BayesianSummarization,Tsvigun2022ActiveSummarization}, we evaluate the performance of the different strategies using the following common AL experimental framework. In each experiment, we assume an annotation budget $B=150$ samples, which are acquired over 15 AL iterations with $s=10$ new samples selected per iteration. At each iteration, the newly annotated samples are then added to the labeled dataset. Following sample selection, we fine-tune a model from its initial pre-trained state on the labeled training set and evaluate its performance on a held-out test set. This evaluation step is not part of the real-world AL process. Rather, it serves the evaluation of the different strategies. In a real-world AL scenario, such extensive evaluation would not be feasible due to the absence of a large labeled test set. This experimental design allows us to observe the impact of different strategies on model performance as the labeled dataset gradually expands. Also note that in a real-world AL scenario, a human annotator would typically label the selected documents at each iteration. However, in our experiments we simulate this by ``revealing" the reference summaries included in the dataset for the samples in the labeled pool. 

% We use the ROUGE-1 score to evaluate the quality of the generated summaries. We focus on ROUGE-1 scores as they provide a good overall indication of summary quality and because we found that other ROUGE metrics follow similar trends. 

We use ROUGE metrics \citep{Lin2004Rouge:Summaries} to assess the quality of the generated summaries, as they offer a reliable measure of summary quality and enable comparison with \cite{Gidiotis2023BayesianSummarization, Tsvigun2022ActiveSummarization}. While we computed ROUGE-1, ROUGE-2 and ROUGE-L metrics, we focus on ROUGE-1 because it provides a good overall indication of summary quality. Furthermore, we noticed that the values of the other metrics follow the same patterns. The complete results, including all ROUGE metrics at key AL steps, are provided in Appendix \ref{appendix-al-iterations}.

Due to the stochastic nature of both the random sampling step and MC dropout and in order to ensure robust results, we repeat each experiment 6 times with different random seeds. In all comparisons, we report the mean performance across all 6 runs, along with the standard deviation to indicate the variability of the results.

\subsection{Models and Training}

To evaluate our AL approach, we conducted experiments on the following Transformer-based  sequence-to-sequence models that have been proven to be effective for summarization tasks:

\begin{itemize}
    \item BART\footnote{\href{https://huggingface.co/facebook/bart-base}{https://huggingface.co/facebook/bart-base}} \cite{Lewis2019BART:Comprehension} is pre-trained as a denoising autoencoder, where the input is corrupted and the model learns to denoise it. This type of pre-training makes it suitable for language generation tasks and a strong performer on a number of NLP tasks including summarization.  The encoder and decoder in the BART\textsubscript{BASE} architecture are built of 6 Transformer blocks.
    \item PEGASUS\footnote{\href{https://huggingface.co/google/pegasus-large}{https://huggingface.co/google/pegasus-large}} \cite{Zhang2020PEGASUS:Summarization} is pre-trained on the C4 \cite{Raffel2020ExploringTransformer} and HugeNews \cite{Zhang2020PEGASUS:Summarization} datasets at a sentence infilling task and achieves strong performance on multiple well established summarization benchmarks. The model's encoder and decoder are built of 16 Transformer blocks each.
    \item FLAN-T5\footnote{\href{https://huggingface.co/google/flan-t5-large}{https://huggingface.co/google/flan-t5-large}} \citep{JMLR:v25:23-0870} adds instruction fine-tuning to the T5 model \cite{Raffel2020ExploringTransformer}, a model with 24 Transformer blocks in both the encoder and decoder. Finetuning leads to improved performance across a wide range of natural language generation tasks \citep{pmlr-v202-longpre23a}, including summarization. When using this model we add the instruction prefix ``Summarize:~" at the beginning of each input. 
\end{itemize}

We used the open-source pre-trained versions of these models from the Hugging Face models repository. To select appropriate training hyperparameters for each model, we conducted a series of fine-tuning experiments on XSUM \citep{xsum-emnlp}, which was not used in our main experiments. After exploring various configurations, we selected a single set of hyperparameters per model that balanced performance and computational efficiency. The full set of hyperparameters can be found in Appendix \ref{appendix-model-hyperparameters}. All experiments were conducted on a single NVIDIA GeForce RTX 3090 GPU.

\subsection{Datasets}

We run experiments on the following summarization datasets: 
\begin{itemize}
    \item AESLC\footnote{\href{https://huggingface.co/datasets/Yale-LILY/aeslc}{https://huggingface.co/datasets/Yale-LILY/aeslc}} \cite{Zhang2020ThisGeneration} consists of 18k emails from the Enron corpus \cite{Klimt2004TheResearch}. The body of each email is used as source text and the subject as summary.
    \item Reddit TIFU\footnote{\href{https://huggingface.co/datasets/ctr4si/reddit\_tifu}{https://huggingface.co/datasets/ctr4si/reddit\_tifu}}  \citep{kim-etal-2019-abstractive} is a dataset sourced from posts on the TIFU subreddit, where users share personal stories. Each story is accompanied by a user-provided "TL;DR" summary. We use the “long” subset, containing around 80k samples.

% is a dataset derived from Reddit posts in the TIFU sub-reddit, where users share personal stories. We use the "long" subset, which contains approximately 80k samples, where the reference summary is a "TL;DR" summary provided by the users.
    \item WikiHow\footnote{\href{https://huggingface.co/datasets/wangwilliamyang/wikihow}{https://huggingface.co/datasets/wangwilliamyang/wikihow}} \citep{koupaee2018wikihowlargescaletext} consists of approximately 200k how-to articles from the wikihow.com website. Each article is divided into multiple paragraphs, with the first sentence of each paragraph serving as the summary. For our task we use the concatenation of these first sentences to create the reference summary for the entire article, similar to the approach followed by \cite{Zhang2020PEGASUS:Summarization}. We evaluate our models on 1.5k out of the 6k samples from the test set to lower the computational complexity.
    \item BillSum\footnote{\href{https://huggingface.co/datasets/FiscalNote/billsum}{https://huggingface.co/datasets/FiscalNote/billsum}} \citep{koupaee2018wikihowlargescaletext,kornilova-eidelman-2019-billsum} is a dataset of approximately 23k US Congressional bills along with their corresponding human-written summaries. Following \cite{Zhang2020PEGASUS:Summarization}, we ignore the California test set and we evaluate our models on 1k out of the 3k samples from the test set to lower the computational complexity.

\end{itemize}

We purposely selected these four datasets to cover a diverse range of domains, summarization styles and document/summary lengths. Dataset details are shown in Table \ref{table:datasets}. We used the publicly available versions of these datasets from the Hugging Face datasets repository.

\begin{table}[h]
\caption{Summarization datasets details}\label{table:datasets}%
\begin{tabular}{@{}llrrr@{}}
\toprule
& & & \multicolumn{2}{c}{Length in words} \\ 
Dataset & Domain  & Dataset size & Source  & Summary  \\
\midrule
AESLC & email & 18,302  & 75 & 5 \\
Reddit TIFU & narrative & 120,000  & 432 & 23 \\
WikiHow & article & 230,000 & 580 & 62 \\
BillSum & legislative text & 22,218 & 1382 & 200 \\
\botrule
\end{tabular}
\end{table}

% \footnote{\url{https://huggingface.co/models}}

\subsection{Embeddings}

Embeddings play a crucial role in similarity-based AL strategies, forming the semantic representations of source documents. Selecting and adapting appropriate embeddings is therefore key for the effectiveness of this kind of methods. In our experiments, we utilize the uncased BERT\textsubscript{BASE}\footnote{\href{https://huggingface.co/google-bert/bert-base-uncased}{https://huggingface.co/google-bert/bert-base-uncased}} \citep{devlin-etal-2019-bert} model to generate embeddings. The dimensionality of the embedding vectors is 768. In order to extract the embedding for each document, we rely on the pooled \texttt{[CLS]} token of the BERT model.

To better align these embeddings with the characteristics of each target domain, we employ the Transformer-based Sequential Denoising AutoEncoder (TSDAE) method \cite{wang-etal-2021-tsdae-using} for unsupervised domain adaptation. TSDAE trains an encoder-decoder architecture where input sentences are intentionally corrupted, and the model learns to reconstruct the original input from the encoded representations, thereby forcing the encoder to capture robust semantic features in its embeddings. By fine-tuning the pre-trained BERT model on unlabeled domain-specific data using TSDAE, we enhance its ability to produce embeddings that reflect domain-specific language usage and semantics. In contrast, \cite{Tsvigun2022ActiveSummarization} uses Task-Adaptive Pre-Training (TAPT) \citep{gururangan-etal-2020-dont}, which continues pre-training the language model on unlabeled domain data using the masked language modeling objective. However, the embeddings generated through TAPT performed worse in our experiments compared to those produced by TSDAE, so we decided to use the latter.

Finally, we evaluated different state-of-the-art embedding models, such as MPNet \cite{Song2020MPNet:Understanding}, without any domain adaptation. Despite the more advanced architectures and strong theoretical performance, we found that the domain-adapted embeddings produced by TSDAE outperformed them in all of our tasks. This outcome underscores the importance of domain adaptation over more advanced embedding models.

% \subsection{Evaluation Metrics}

% We use the ROUGE-1 score to evaluate the quality of the generated summaries. We focus on ROUGE-1 scores as they provide a good overall indication of summary quality and because we found that other ROUGE metrics follow similar trends. 

% \subsubsection{This is an example for third level head---subsubsection head}

% Sample body text. Sample body text. Sample body text. Sample body text. Sample body text. Sample body text. Sample body text. Sample body text. 

\section{Results and Discussion}\label{sec:results}
This section presents the results of the experiments we conducted on 4 benchmark datasets with 3 different summarization models. First, we show the learning curves of DUAL and compare them with different diversity and uncertainty based strategies. Second, we present a visual analysis of the embedding space, showcasing the differences in the way these methods select samples. Last but not least, we focus on the diversity of the selected data, and discuss the topic of outliers.

\subsection{Performance}

\begin{figure}[h]
\centering
\hfill
\includegraphics[width=0.99\textwidth]{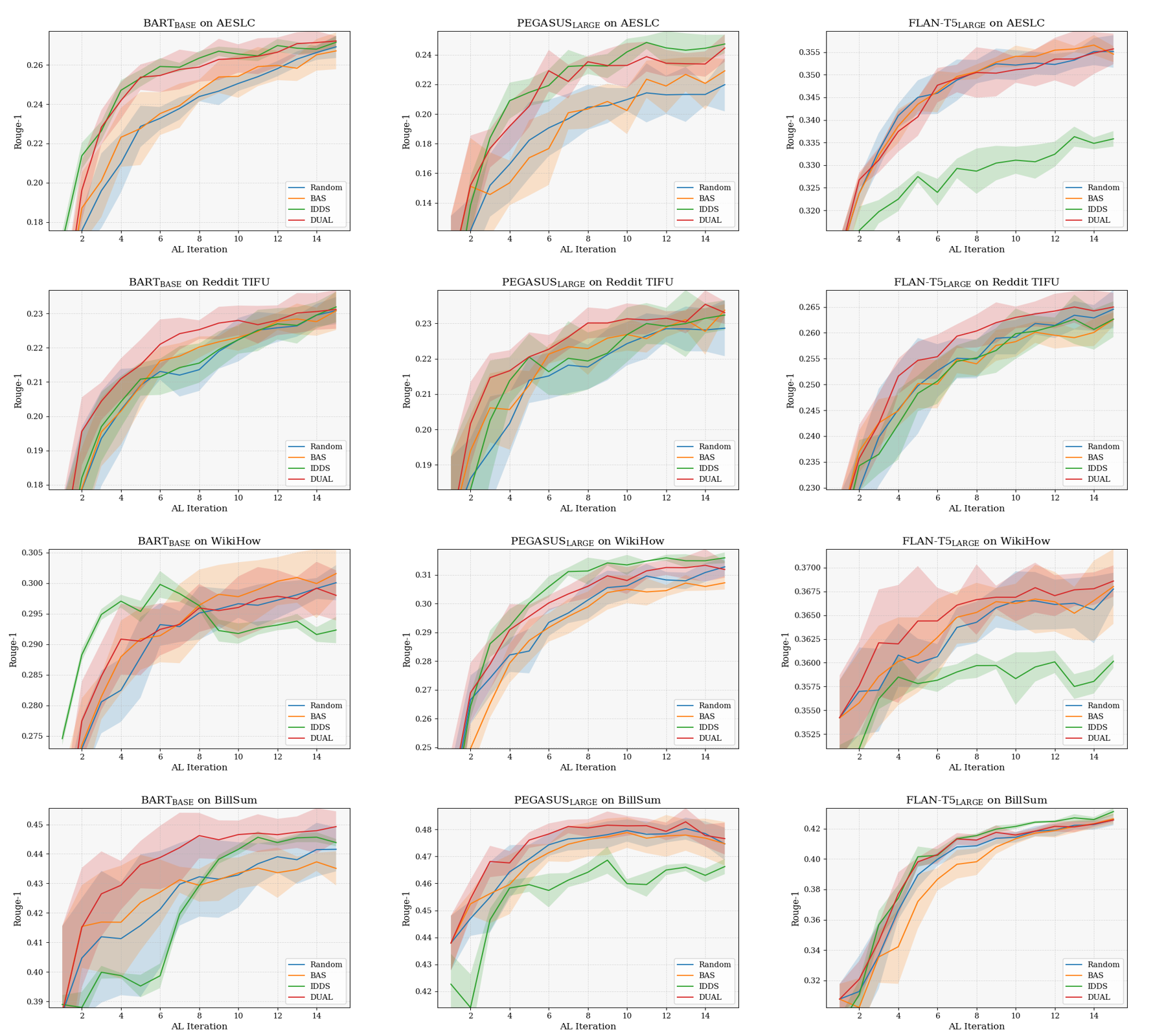}
\caption{Rouge-1 scores of AL strategies across models and datasets}
\label{fig:combined_results}
\end{figure}

Figure \ref{fig:combined_results} illustrates the performance of all AL strategies across different models for each of our experimental datasets. We obtain the learning curves by computing ROUGE-1 scores on the test set at the end of every AL iteration. The lines represent the mean, while the shaded areas indicate the standard deviation across 6 runs with different random seeds. Table \ref{table:rouge-al-iterations-all} in Appendix \ref{appendix-al-iterations} provides a detailed comparison of ROUGE-1, ROUGE-2, and ROUGE-L scores at key AL iterations (5, 10, 15) for all datasets, models, and strategies.

%The results of our experiments demonstrate that DUAL generally outperforms all other strategies across most model and dataset combinations. Even when it doesn't show a clear advantage, DUAL closely follows the best performing model.

% Observation 1: High variability on same dataset but different models %
We observe that the performance of AL strategies on a particular dataset can vary significantly between different models. In WikiHow, BAS performs best for the BART\textsubscript{BASE} model, IDDS excels for the PEGASUS\textsubscript{LARGE} model, and DUAL achieves the best results for FLAN-T5\textsubscript{LARGE}. This variability suggests that the behavior of uncertainty and diversity as AL strategies is not universal across all models, and highlights the challenge of finding an AL strategy that is effective across all different models. Nonetheless, DUAL demonstrates consistently good performance across all models and dataset combinations, showcasing competitive results in all different scenarios.

Another notable observation is that in certain experimental setups, such as FLAN-T5\textsubscript{LARGE} on Reddit TIFU and PEGASUS\textsubscript{LARGE} on BillSum, both BAS and IDDS fail to match the performance of random selection. This further highlights the inconsistency of AL strategies in summarization tasks \cite{Tsvigun2022ActiveSummarization,Gidiotis2023BayesianSummarization}. In contrast, DUAL consistently outperforms or matches the performance of random selection in all experimental setups, emerging as a more reliable approach. This consistent performance can be attributed to DUAL's design, which integrates both uncertainty and diversity based characteristics, as well as the inclusion of random sampling as a core component of the method.

We now focus on the combinations of datasets and models where the best performance is achieved, as these are the cases with higher practical value. We notice that FLAN-T5\textsubscript{LARGE} achieves the best results in AESLC, Reddit TIFU and WikiHow, while PEGASUS\textsubscript{LARGE} achieves the best results in BillSum. There are several considerations that align with these observations. First, FLAN-T5\textsubscript{LARGE} has 770 million parameters, significantly larger than PEGASUS\textsubscript{LARGE} (568M) and BART\textsubscript{BASE} (139M). Consequently, it is expected to generalize better across different formats, styles, and text lengths. Second, FLAN-T5\textsubscript{LARGE} is trained on the full C4 dataset using a text-to-text approach, enhancing its adaptability. PEGASUS\textsubscript{LARGE} also leverages C4 but only partially, with a primary focus on HugeNews. As a result, PEGASUS\textsubscript{LARGE} is more specialized for document-level summarization, explaining its superior performance on BillSum. Interestingly in 3 out of these 4 cases (all except Reddit TIFU) IDDS achieves significantly worse results than the rest of the methods. 

% Observation: Sampling data from the same region 
With BART\textsubscript{BASE} on WikiHow, IDDS shows a strong start, outperforming other strategies in the early stages of learning. However, its performance begins to decline after about 40-50 samples, eventually falling below random sampling and other strategies. Notably, the performance drop in IDDS is not caused by random fluctuations. Rather, it stems from the limitations of the sample selection strategy, as IDDS consistently chooses the same set regardless of the training seed. In contrast, DUAL demonstrates more stable performance throughout the AL process, as it avoids the sharp decline seen in IDDS.

\subsection{Visualization Analysis of Selected Samples}

% To gain deeper insights into the behavior of our proposed strategy and compare it with other approaches, we conducted a visualization analysis of the distribution of the samples in the embeddings space, as selected by different active learning strategies. For each experiment, we chose one seed out of the six that were executed. We plot the samples chosen by each active learning strategy alongside all other samples in the dataset. To facilitate visualization, we first reduced the embedding dimensions to two using PCA. 

\begin{figure}[h]
\centering
\hfill
\includegraphics[width=0.99\textwidth]{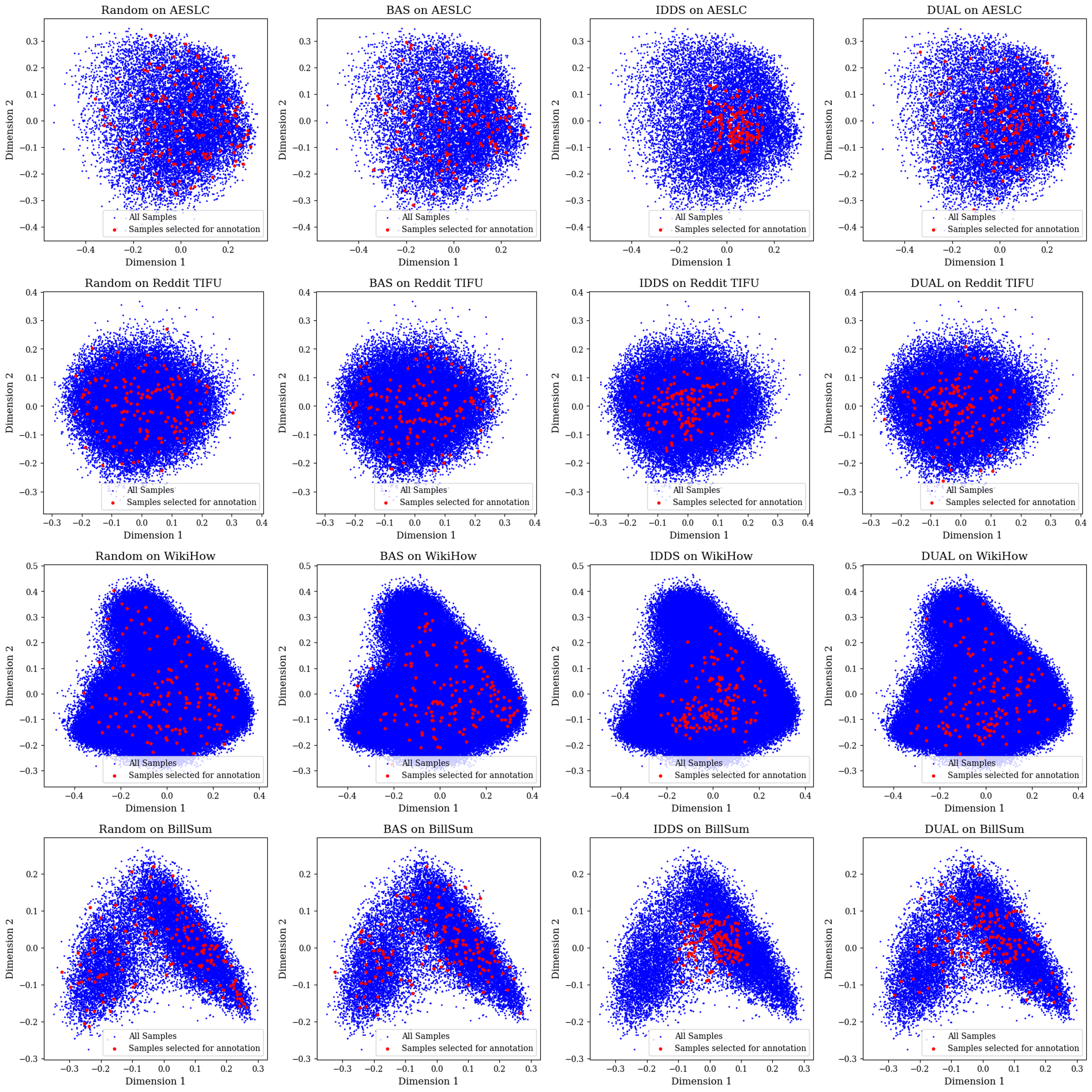}
\caption{Visualization of the embedding space showing selected samples by each AL strategy in each dataset}
\label{fig:combined_embeddings}
\end{figure}

In order to understand the differences between selection strategies, we present a visual analysis of the sample distributions selected by each strategy. For all datasets, we apply PCA to reduce the dimensionality of the BERT embedding vectors to two dimensions for visualization. Figure \ref{fig:combined_embeddings} shows the samples that were selected for labeling by each AL strategy. Since each experiment was repeated multiple times with different models and random seeds, the selected samples vary slightly across different experimental runs, but the distributions are similar.

The plotted distributions reveal that IDDS can become confined within a certain region of the embedding space. As we have discussed earlier, this behavior can be problematic for several reasons. %As shown in Figure \ref{fig:combined_embeddings}, IDDS embeddings tend to cluster in dense areas of the dataset. 
The lack of exploration prevents the model from being trained on samples that are not represented in those regions. This limitation becomes even more pronounced because the IDDS distribution is only based on the initial embeddings and will never change throughout the AL training. This effectively means that if IDDS gets stuck in a region that's not representative enough, there's no way to explore samples outside this region. Furthermore, in cases where the data distribution includes more than one cluster, such as in Wikihow and BillSum, IDDS tends to focus on one cluster, missing other parts of the data space, or can even select outliers between the two clusters (BillSum). This will inevitably lead to a training set that is not indicative of the real data distribution.

In contrast, we observe that both random sampling and BAS select samples that cover the entire data distribution. Unfortunately, both of these methods seem to also select a relatively high amount of outlier samples, which is likely to harm summarization performance.

DUAL on the other hand avoids both of the aforementioned issues, as it samples with high probability from the denser regions of the space, but also explores other regions. As we can see in the Wikihow and BillSum results, unlike IDDS, DUAL selects data samples from both clusters of the data space. In addition, the majority of samples selected by DUAL come from the dense parts of the data distribution with much fewer outliers compared to random and BAS. As a result, the training set created by DUAL exhibits a good coverage of the entire data space.

% The visualization also shows that random sampling covers a broader range of the distribution. However, it still selects a relatively high amount of outlier samples, which may not always be representative or informative for the training process. In contrast, IDDSBAS strikes a balance between outlier selection and diversity. It avoids being restricted to dense regions of the embedding space and instead covers a more uniform distribution of selected samples, while also choosing high uncertainty samples.

Finally, this analysis highlights the significance of embedding pre-training in the performance of methods that rely on diversity. The quality of the embeddings directly impacts the ability of both DUAL and IDDS to effectively select representative samples. As explained, this limitation is somewhat but not completely mitigated in DUAL by incorporating random sampling in the process.

\subsection{Diversity and Outliers}\label{}

% Building upon the insights gained from the previous section, w
Driven by the findings of the embeddings analysis, we further investigated the behavior of different AL strategies with regards to diversity and outliers. To quantify these aspects, we rely on the two additional metrics defined on the labeled set $L$. Following \cite{perlitz-etal-2023-active}, we measure the diversity score of $L$ by computing the average Euclidean distance of all samples from the center, using the BERT embeddings from our previous experiments. In addition, we compute the outlier score based on the KNN-density proposed in \cite{zhu-etal-2008-active}. More specifically, we quantify density by taking the average Euclidean distance between an instance and its $K=10$ nearest neighbors within $U$. The outlier score of $L$ is then defined as the average density of its instances, where a higher density corresponds to a lower outlier score.

\begin{figure}[h]
\centering
\hfill
\includegraphics[width=0.99\textwidth]{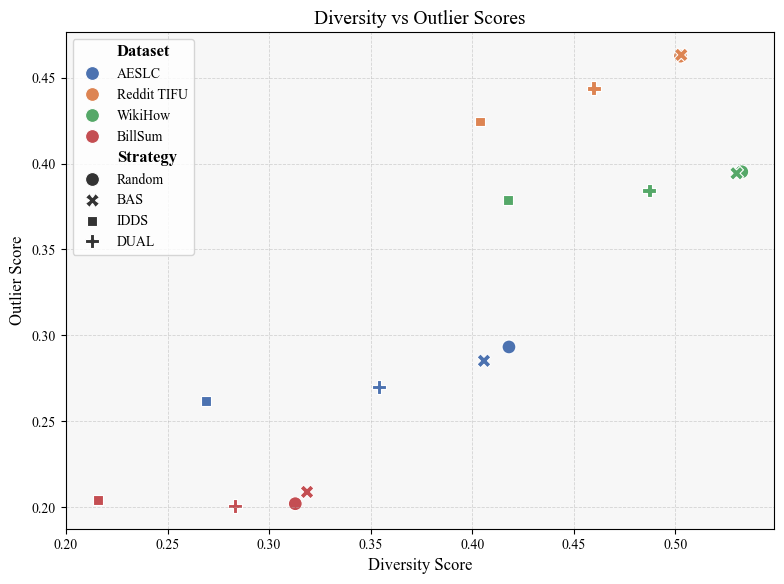}
\caption{Diversity and outlier scores for different AL strategies in each dataset}
\label{fig:diversity_and_outlier}
\end{figure}

Figure \ref{fig:diversity_and_outlier} shows the diversity and outlier scores for different AL strategies per dataset. On one hand, IDDS consistently shows the lowest outlier scores but at the same time exhibits lower diversity scores, reflecting its tendency to select samples from the denser regions of the embedding space. On the other hand, BAS and Random show the highest diversity, but this comes at the cost of higher outlier scores, particularly in AESLC and Reddit TIFU. This suggests that they are more prone to selecting atypical or noisy examples that can potentially harm summarization performance. DUAL strikes a balance, consistently achieving good diversity and low outliers across all datasets. In other words, it sufficiently explores the data space, while also avoiding most of the outliers.

% \begin{figure}[h]
% \centering
% \hfill
% \includegraphics[width=0.99\textwidth]{images/diversity-outlier-scores.png}
% \caption{Main caption for both images}
% \label{fig:Diversity and Outlier scores}
% \end{figure}

\subsection{Computational Efficiency}\label{}

While computational efficiency is not the primary focus of AL, it remains an important aspect to consider in practical applications. Uncertainty-based approaches generally require forward-passing each candidate sample through the model to estimate uncertainty. Specifically, BAS leverages multiple stochastic forward passes via Monte Carlo dropout, magnifying the computational overhead. Since these forward passes are performed at each AL iteration to reassess the uncertainty of the unlabeled pool or a portion of it, the cost can grow considerably with larger datasets and bigger models. Diversity-based strategies typically do not utilize the model during the AL iterations, as they primarily operate on pre-computed embeddings. However, these strategies still require an upfront computational cost due to the embedding computation and potential domain adaptation. Again, this cost can be substantial, especially for large datasets.

DUAL combines both uncertainty and diversity, thus incurring the respective costs of each component. In practice, however, incorporating the random sampling step and using uncertainty estimates only on smaller subsets (i.e. the top-$k$ IDDS candidates) can help contain computational demands. This means that in many real-world scenarios, DUAL’s overhead remains manageable. Embeddings are computed once and reused, and Monte Carlo dropout inference is restricted to a narrower set of promising samples.

% Uncertainty-based strategies require an ``iterative learning" approach, where samples must be acquired sequentially. For each AL iteration, we need to train the model on the current labeled set to make informed decisions about the next batch of samples to select. This iterative training process may significantly increases the overall computational cost. Furthermore, BAS is inherently more computationally expensive due to the need for multiple forward passes through the model for each sample being evaluated. 

% In contrast, diversity-based strategies like IDDS do not require this iterative model training, as they primarily operate on pre-computed embeddings. However, these strategies still require an upfront computational cost due to the embedding computation and potential domain adaptation. This cost can be substantial, especially for large datasets. Consequently, since DUAL combines both uncertainty and diversity-based approaches, its computational complexity is governed by both factors discussed above.

Table \ref{table:computation_times} presents average sample selection times per AL iteration for both BAS and DUAL strategies across various datasets and models, alongside the corresponding training times. Note that we do not show the sample selection times for Random and IDDS because their selection overhead is 0. These results demonstrate that DUAL consistently requires less time for sample selection than BAS. For example, with BART$_{\text{BASE}}$ on AESLC, DUAL takes about 20.71 seconds compared to BAS’s 24.86 seconds, and on Reddit TIFU with FLAN-T5$_{\text{LARGE}}$, the selection time is reduced from 177.64 seconds (BAS) to 110.50 seconds (DUAL). Note that DUAL still requires the computation of embeddings for all samples as a preliminary step in its selection process. The training time is included solely to provide context regarding the overall duration of the AL cycles and does not depend on the specific AL strategy employed.

\begin{table}[ht]
\centering
\caption{Average sample selection times (in seconds) per AL iteration for BAS and DUAL, and training times (in seconds) for each dataset--model combination. The average train time per AL step is roughly the same for all strategies since each AL iteration uses the same amount of data. For other strategies (Random, etc.), the average sample selection duration is 0.}
\label{table:computation_times}
\begin{tabular}{llrrr}
\toprule
 &  & \multicolumn{2}{c}{\textbf{Avg. Selection Time}} &  \\
\cmidrule(lr){3-4}
\textbf{Dataset} & \textbf{Model} & \textbf{BAS} & \textbf{DUAL} &  \textbf{Train Time}\\
\midrule
\multirow{3}{*}{\textbf{AESLC}} 
    & BART\textsubscript{BASE} & 24.86 & 20.71 & 74.07 \\
    & PEGASUS\textsubscript{LARGE} & 46.57 & 38.57 & 102.20 \\
    & FLAN-T5\textsubscript{LARGE} & 81.71 & 54.71 & 27.93 \\
\midrule
\multirow{3}{*}{\textbf{Reddit TIFU}}
    & BART\textsubscript{BASE} & 23.93 & 19.93 & 82.80 \\
    & PEGASUS\textsubscript{LARGE} & 75.93 & 53.43 & 113.53 \\
    & FLAN-T5\textsubscript{LARGE} & 177.64 & 110.50 & 40.07 \\
\midrule
\multirow{3}{*}{\textbf{WikiHow}}
    & BART\textsubscript{BASE} & 25.12 & 20.57 & 106.73 \\
    & PEGASUS\textsubscript{LARGE} & 85.71 & 59.93 & 133.00 \\
    & FLAN-T5\textsubscript{LARGE} & 121.36 & 87.29 & 147.40 \\ % EXPLAIN HERE
\midrule
\multirow{3}{*}{\textbf{BillSum}}
    & BART\textsubscript{BASE} & 36.43 & 31.07 & 116.13 \\
    & PEGASUS\textsubscript{LARGE} & 149.40 & 110.12 & 124.60 \\
    & FLAN-T5\textsubscript{LARGE} & 172.64 & 128.00 & 46.20 \\
\bottomrule
\end{tabular}
\end{table}

\section{Conclusion}
\label{sec:conclusion}
This work demonstrates how uncertainty and diversity based strategies can be combined in an AL setting. Extensive experiments across several benchmark datasets, and using various popular models, show that the combination consistently outperforms each individual strategy as well as random sampling. Our analysis indicates that this is because DUAL successfully addresses the limitations of BAS and IDDS, such as noisy samples and lack of exploration.

In addition, we study the effect of different strategies with regards to the selected data distribution and outliers. By visualizing the distribution of selected data points and analyzing diversity, we get a better understanding of the common issues each strategy faces. These insights allow us to develop several techniques, such as including a percentage of random samples and discarding candidate IDDS samples when they are not selected, in order to safeguard against those issues.

\subsection{Future Work}
While the proposed hybrid AL framework effectively combines uncertainty and diversity, there are several promising avenues for future exploration. 

First, while BLEU variance (BLEUVar) and other traditional metrics have been effective for uncertainty estimation, their task-agnostic nature may limit their applicability in certain summarization tasks. Incorporating task-specific metrics, such as content preservation or factual consistency, could provide a more robust basis for evaluating uncertainty and selecting data instances.

Second, the current framework was evaluated on benchmark datasets, which may not fully represent the challenges posed by large-scale, real-world datasets. Future research could focus on scaling the framework to handle extensive datasets while maintaining computational efficiency. This might involve techniques such as model distillation or sparse sampling methods.

Finally, incorporating human feedback as part of the AL loop could further refine the hybrid framework. For example, human evaluators could help validate or adjust the sampled instances, especially in cases where the model's uncertainty estimates may not align with human judgment.

By addressing these directions, future research can not only refine the current hybrid framework but also advance the broader field of active learning for text summarization. Continued exploration of this topic has the potential to unlock even greater efficiency and effectiveness in summarization tasks across diverse real-world applications.

\bibliography{sn-bibliography,references}

\newpage

\begin{appendices}

% \section{Section title of first appendix}\label{secA1}

\section{Hyperparameters}\label{appendix-model-hyperparameters}

\begin{table}[h]
\caption{Train hyperparameters}\label{tab1}%
\begin{tabular}{@{}llll@{}}
\toprule
Hyperparameter & BART\textsubscript{BASE}  & PEGASUS\textsubscript{LARGE} & FLAN-T5\textsubscript{LARGE}\\
\midrule
Number of train epochs   & -   & -  & 3  \\
Train batch size    &  16  & 8  & 6\footnotemark[1]  \\
Min. train steps    & 350   & 200  & -  \\
Optimizer    & AdamW  & AdamW  & Adafactor \\
Learning rate    & 2e-5    & 5e-4  & 3e-5  \\
Weight decay    & 0.028   & 0.03  & 0.01  \\
% Max gradient norm    & 0.28   & data 5  & 1.0  \\
Warmup ratio   & 0.1   & 0.1  & 0.1  \\
Num. of beams at evaluation   & 4   & 4 & 3  \\
\botrule
\end{tabular}
% \footnotetext{Source: This is an example of table footnote. This is an example of table footnote.}
\footnotetext[1]{We set the train batch size to 4 for the WikiHow and BillSum datasets due to memory constraints.}

\end{table}

\section{Performance of different AL iterations}\label{appendix-al-iterations}

\begin{table}[htbp]
\centering
{\fontsize{8}{8}\selectfont
\caption{Rouge scores (R1/R2/RL) at key AL iterations (5, 10, 15) across all datasets, strategies, and models. Bold indicates the highest score for each experiment.}
\label{table:rouge-al-iterations-all}
\begin{tabular}{llccc}
\toprule
\textbf{Model} & \textbf{Dataset / Strategy} & \textbf{Iter. 5} & \textbf{Iter. 10} & \textbf{Iter. 15} \\
\midrule
\multicolumn{5}{c}{\(\text{BART}_\text{BASE}\)} \\
\midrule
AESLC  & Random  & 22.87/11.23/22.42 & 25.05/12.49/24.50 & 26.92/13.37/26.19 \\
       & BAS     & 22.76/11.26/22.33 & 25.41/12.68/24.81 & 26.71/13.22/26.06 \\
       & IDDS    & 25.32/\textbf{12.82}/\textbf{24.94} & \textbf{26.55}/13.18/\textbf{25.91} & 27.15/13.62/26.48 \\
       & DUAL    & \textbf{25.37}/12.75/24.88 & 26.32/\textbf{13.24}/25.72 & \textbf{27.21}/\textbf{13.62}/\textbf{26.50} \\
\cmidrule(lr){2-5}
Reddit TIFU & Random  & 20.90/4.81/16.29 & 22.24/5.40/17.41 & 23.09/5.78/18.00 \\
       & BAS     & 20.88/4.60/16.06 & 22.31/5.38/17.38 & 23.08/\textbf{5.81}/18.06 \\
       & IDDS    & 21.08/4.77/16.28 & 22.24/5.24/17.21 & \textbf{23.19}/5.77/\textbf{18.11} \\
       & DUAL    & \textbf{21.50}/\textbf{4.91}/\textbf{16.68} & \textbf{22.80}/\textbf{5.51}/\textbf{17.68} & 23.12/5.79/18.02 \\
\cmidrule(lr){2-5}
WikiHow & Random  & 28.79/8.12/20.30 & 29.66/8.94/21.34 & 30.01/9.44/\textbf{21.84} \\
       & BAS     & 29.09/8.28/20.29 & \textbf{29.78}/\textbf{8.95}/21.29 & \textbf{30.16}/\textbf{9.48}/21.80 \\
       & IDDS    & \textbf{29.54}/\textbf{8.56}/\textbf{20.87} & 29.18/8.95/\textbf{21.51} & 29.23/9.36/21.82 \\
       & DUAL    & 29.05/8.24/20.31 & 29.60/8.91/21.26 & 29.80/9.38/21.77 \\
\cmidrule(lr){2-5}
BillSum & Random  & 41.57/\textbf{21.15}/29.50 & 43.29/22.29/30.76 & 44.16/22.85/\textbf{31.32} \\
       & BAS     & 42.35/21.08/29.64 & 43.35/22.17/30.71 & 43.51/22.74/31.12 \\
       & IDDS    & 39.52/20.29/28.51 & 44.15/21.57/30.26 & 44.38/22.40/31.00 \\
       & DUAL    & \textbf{43.64}/21.12/\textbf{29.76} & \textbf{44.66}/\textbf{22.31}/\textbf{30.87} & \textbf{44.93}/\textbf{22.86}/31.30 \\
\midrule
\multicolumn{5}{c}{\(\text{PEGASUS}_\text{LARGE}\)} \\
\midrule
AESLC  & Random  & 18.48/8.31/18.01 & 20.84/9.69/20.34 & 22.68/10.61/22.12 \\
       & BAS     & 18.40/7.79/18.00 & 21.11/9.64/20.51 & 22.02/9.82/21.39 \\
       & IDDS    & \textbf{22.21}/\textbf{10.20}/\textbf{21.68} & \textbf{23.52}/\textbf{10.64}/\textbf{23.05} & \textbf{24.74}/11.28/\textbf{23.94} \\
       & DUAL    & 20.01/9.27/19.48 & 22.00/10.19/21.45 & 24.34/\textbf{11.51}/23.63 \\
\cmidrule(lr){2-5}
Reddit TIFU & Random  & 21.39/4.71/16.50 & 22.43/5.21/17.19 & 22.86/5.41/17.60 \\
       & BAS     & 21.25/4.61/16.29 & 22.68/5.27/17.34 & \textbf{23.37}/5.54/\textbf{17.96} \\
       & IDDS    & 22.05/\textbf{5.08}/\textbf{16.81} & 22.68/5.36/17.35 & 23.23/5.52/17.69 \\
       & DUAL    & \textbf{22.05}/4.93/16.79 & \textbf{23.13}/\textbf{5.40}/\textbf{17.74} & 23.30/\textbf{5.57}/17.77 \\
\cmidrule(lr){2-5}
WikiHow & Random  & 28.35/8.21/20.75 & 30.62/9.50/22.62 & 31.28/10.01/23.30 \\
       & BAS     & 28.72/8.13/20.73 & 30.50/9.21/22.20 & 30.73/9.44/22.47 \\
       & IDDS    & \textbf{30.97}/\textbf{9.47}/\textbf{22.50} & \textbf{31.29}/\textbf{10.05}/\textbf{23.38} & \textbf{31.58}/\textbf{10.30}/\textbf{23.72} \\
       & DUAL    & 29.29/8.52/21.36 & 30.92/9.46/22.50 & 31.47/10.09/23.43 \\
\cmidrule(lr){2-5}
BillSum & Random  & 46.90/24.77/32.83 & 47.96/26.31/\textbf{34.25} & 47.46/\textbf{26.81}/\textbf{34.67} \\
       & BAS     & 46.74/24.75/32.74 & 47.89/25.98/33.91 & 47.47/26.72/34.60 \\
       & IDDS    & 45.96/24.61/32.36 & 45.99/25.06/32.58 & 46.62/26.02/33.79 \\
       & DUAL    & \textbf{47.61}/\textbf{25.26}/\textbf{33.16} & \textbf{48.13}/\textbf{26.40}/34.08 & \textbf{47.66}/26.69/34.56 \\
\midrule
\multicolumn{5}{c}{\(\text{FLAN-T5}_\text{LARGE}\)} \\
\midrule
AESLC  & Random  & \textbf{34.50}/\textbf{18.91}/\textbf{33.19} & 35.21/19.37/34.03 & 35.51/\textbf{19.59}/\textbf{34.38} \\
       & BAS     & 34.35/18.68/33.04 & \textbf{35.41}/\textbf{19.53}/\textbf{34.22} & 35.46/19.58/34.29 \\
       & IDDS    & 32.75/17.71/31.43 & 33.11/17.75/31.75 & 33.58/18.25/32.26 \\
       & DUAL    & 34.07/18.51/32.74 & 35.11/19.24/33.71 & \textbf{35.57}/19.43/34.10 \\
\cmidrule(lr){2-5}
Reddit TIFU & Random  & 24.97/7.83/19.97 & 25.92/8.22/20.79 & 26.46/\textbf{8.44}/21.19 \\
       & BAS     & 25.02/7.87/20.01 & 25.83/8.23/20.67 & 26.26/8.34/21.04 \\
       & IDDS    & 24.83/7.78/19.74 & 25.98/8.18/20.71 & 26.26/8.34/20.95 \\
       & DUAL    & \textbf{25.47}/\textbf{8.02}/\textbf{20.31} & \textbf{26.30}/\textbf{8.29}/\textbf{21.02} & \textbf{26.50}/8.39/\textbf{21.20} \\
\cmidrule(lr){2-5}
WikiHow & Random  & 36.00/15.90/29.85 & 36.65/16.04/30.18 & 36.77/\textbf{16.09}/\textbf{30.25} \\
       & BAS     & 36.08/15.95/29.90 & 36.62/16.07/30.18 & 36.80/16.05/30.19 \\
       & IDDS    & 35.78/15.83/29.66 & 35.83/15.93/29.81 & 36.02/15.92/29.90 \\
       & DUAL    & \textbf{36.44}/\textbf{16.17}/\textbf{30.28} & \textbf{36.69}/\textbf{16.14}/\textbf{30.29} & \textbf{36.86}/16.01/30.14 \\
\cmidrule(lr){2-5}
BillSum & Random  & 38.95/20.86/29.97 & 41.43/22.10/31.82 & 42.58/22.94/32.65 \\
       & BAS     & 37.20/20.42/28.70 & 41.33/\textbf{22.91}/\textbf{32.05} & 42.65/\textbf{23.87}/\textbf{33.02} \\
       & IDDS    & \textbf{40.15}/\textbf{21.37}/\textbf{30.20} & \textbf{42.14}/22.22/31.72 & \textbf{43.12}/23.19/32.64 \\
       & DUAL    & 39.81/21.19/30.17 & 41.58/22.25/31.83 & 42.56/23.04/32.51 \\
\bottomrule
\end{tabular}
}
\end{table}

%An appendix contains supplementary information that is not an essential part of the text itself but which may be helpful in providing a more comprehensive understanding of the research problem or it is information that is too cumbersome to be included in the body of the paper.




\end{appendices}



\end{document}